\pdfoutput=1
\documentclass[11pt]{article}

\usepackage[final]{acl}

\usepackage{times}
\usepackage{latexsym}

\usepackage[T1]{fontenc}
\usepackage[utf8]{inputenc}

\usepackage{microtype}

\usepackage{inconsolata}

\usepackage{graphicx}

\usepackage{enumitem}
\usepackage{amsmath,amssymb} % For the checkmark
\usepackage{colortbl}
\usepackage{soul} % To highlight
\usepackage{multirow} % For table design
\usepackage{array, makecell} % For multiple lines in overleaf tables
\usepackage{float}
\usepackage{booktabs}
\usepackage{pifont}
\usepackage{listings}
\usepackage[scaled]{beramono}
\usepackage{siunitx, mhchem}
\usepackage[export]{adjustbox}
\usepackage{xspace}
\usepackage{subfig}
\usepackage{numprint}
\npthousandsep{\,}
\usepackage{tikz}
\usepackage{url}
\usepackage{bbm}

\newcommand{\greencheck}{{\color{teal}\ding{51}}}
\newcommand{\redcross}{{\color{purple}\ding{55}}}

\newcommand{\colormap}[2]{%
    \pgfmathparse{100 - 100 * #1} % Inverte la percentuale (100% giallo per 0, 100% verde per 1)
    \xdef\result{\pgfmathresult} % Salva il risultato in una variabile
    \cellcolor{yellow!\result!green}{#2} % Applica il colore e stampa il punteggio
}

\usepackage{hyperref}

\title{Hatevolution: What Static Benchmarks Don't Tell Us}

\author{
 \textbf{Chiara Di Bonaventura\textsuperscript{1,2}},
 \textbf{Barbara McGillivray\textsuperscript{1}}, 
 \textbf{Yulan He\textsuperscript{1}},
 \textbf{Albert Meroño-Peñuela\textsuperscript{1}}
\\
 \textsuperscript{1}King's College London 
 \textsuperscript{2}Imperial College London 
\\
 \small{
   \href{mailto:email@domain}{chiara.di\_bonaventura@kcl.ac.uk}
 }
}

\begin{document}
\maketitle
\begin{abstract}
Language changes over time, including in the hate speech domain, which evolves quickly following social dynamics and cultural shifts. 
While NLP research has investigated the impact of language evolution on model training and has proposed several solutions for it, its impact on model benchmarking remains under-explored. 
Yet, hate speech benchmarks play a crucial role to ensure model safety. 
In this paper, we empirically evaluate the robustness of 20 language models across two evolving hate speech experiments, and we show the temporal misalignment between static and time-sensitive evaluations. 
Our findings call for time-sensitive linguistic benchmarks in order to correctly and reliably evaluate language models in the hate speech domain.
\end{abstract}

\section{Introduction}

Language continuously evolves adapting to social and cultural dynamics \cite{altmann2009beyond_word_frequency,eisenstein2014lexical_change_social_media,labov2011principles_linguistic_changes}, e.g., words gain new meanings or lose their existing ones, words shift polarity, and new words emerge.
This language evolution challenges NLP models across different domains \cite{alkhalifa2023building_for_tomorrow,luu2022temporal_misalignment}, with hate speech being one of the most challenging due to the semantic broadening of harm-related concepts in the past 50 years \cite{vylomova2021semantic_broadening_harm}, frequent changes in words' polarity \cite{mcgillivray2022time_sensitive_offensive_detection}, and reclaimed language \cite{zsisku2024reclaimed_hate_speech}. 
Indeed, \citet{dibonaventura2025detection_explanation_LLM} recently show that language models' distributional knowledge can be enhanced with temporal linguistic knowledge to effectively detect and explain hateful content.
While NLP research has extensively investigated the impact of hate speech evolution in model training paradigms, showing that temporal misalignment between training and test sets leads to decreasing performance over time (i.e., temporal bias) across models and languages (\citealt{florio2020time_hate,jin2023examining_temporal_bias}, \textit{inter alia}), the implications of evolving hate speech in model benchmarking have not been explored.
%are still under-explored. 

Yet, hate speech benchmarks play a crucial role as they are widely embedded in the safety evaluation of language models (e.g., \citealt{gehman2020realtoxicityprompts,liang2023helm,ying2024safebench}), which are increasingly used in real-world applications and decision making \cite{bavaresco2024llms_human_judges,zheng2023llm_as_a_judge}. 
Although these benchmarks provide a comprehensive comparison of language models that would not be possible with held-out test sets, they face the same issue: they are static. 
In other words, they are grounded to the specific timestamp in which they were developed, and consequently they cannot account for language change. 
We argue that evolving hate speech plays a role in the reliability of static model benchmarking over time, potentially leading to an overestimation of language models' safety in light of well-known issues like temporal bias and benchmark saturation \cite{sainz2023data_contamination,sainz2023chatgpt}.
%data contamination, and data saturation
Therefore, we seek to answer \textit{``how does static hate speech benchmarking correlate with evolving language?''}.

By providing empirical evidence of this temporal challenge in model benchmarking, we hope our study will raise awareness in the risks associated with static evaluations of language models, and will fuel research towards time-sensitive evaluations of NLP models in a similar way in which studies that investigated the impact of language evolution on model training led to the development of alternative solutions, e.g., temporal attention \cite{rosin2022temporal_attention} or the injection of time-sensitive lexical information \cite{mcgillivray2022time_sensitive_offensive_detection}.

\section{Evolving Hate Speech}
To answer our research question, we first design two experiments for evolving hate speech detection accounting for different aspects of language evolution, and we propose two time-sensitive metrics to evaluate language models.
Then, we evaluate the same models on static hate speech benchmarks, and we measure the correlation in models' ranking across time-sensitive and static evaluations.\footnote{The data and code are available at \url{https://github.com/ChiaraDiBonaventura/hatevolution/tree/main}.}

\paragraph{Experiment 1: Time-Sensitive Shifts.}
% time-sensitive language shifts
We investigate contextual evolution of hate speech, focusing on time-sensitive shifts, such as semantic, topical, and polarity changes. 
For instance, the word `gammon' has undergone multiple transformations simultaneously \cite{mcgillivray2022time_sensitive_offensive_detection}: a \emph{semantic change} from referring to food (ham) to a political insult; a \emph{topic shift} towards political discourse; and a \emph{polarity shift} towards negativity. 
In contrast, certain terms targeting Asian communities predominantly experienced a polarity shift, becoming more offensive during the Coronavirus pandemic \cite{huang2023anti_asian_covid}. 
Moreover, time-sensitive shifts might manifest as changes in the cultural perception of what is considered offensive, e.g., reclaimed slurs. 
These time-sensitive shifts are notoriously difficult to disentangle \cite{luu2022temporal_misalignment}, and we do not attempt to do so in this work. 
Instead, we aim to quantify how their complex interplay affects model performance over time, and in turn how this time-sensitive performance correlates with performance on static benchmarks. 
To study this, we use the English version of the Singapore Online Attack dataset \cite{haber2023singapore-attacks} as it has the biggest and most recent coverage of annotated texts with timestamp information for hate speech research (i.e., 2011-2022 Reddit posts). 
% contains hateful and non-hateful Reddit posts with timestamp when they were written, covering 2011-2022. 
%Note that the fact that this dataset is drawn from a single social media platform ensures consistency in the language. 
We evaluate models with \textbf{time-sensitive macro F1} defined as \(\frac{1}{T}\sum_{t=1}^{T} F1_t\), where \(F1_t\) is the macro-averaged F1 specific to year \(t\). 
This allows to measure how well language models adapt to evolving contexts of hate speech due to yearly time-sensitive shifts. 
Ideally, we want language models to exhibit high and stable time-sensitive F1 scores over time. 
We limit the analysis to 2017-2022 as there were not enough data before 2017.

%% extra comments
% Report the averaged difference between f1 in 2022 and f1 in 2017 across models, in absolute value. 
% Should we expect our models to age that much in the next three years (considering that it's 2024 and two years have already passed since 2022, which is the latest year available for evaluation)?

% Nice to investigate: label distribution over time, KL divergence (similar to paper "time waits for no one!")

\paragraph{Experiment 2: Vocabulary Expansion.}
% language expansion
We examine language expansion, focusing on the emergence of neologisms, i.e., newly coined terms that have entered our vocabulary. 
To measure model robustness to this type of language evolution, we extend the NeoBench dataset \cite{zheng2024neobench} to the task of hate speech detection.  
Specifically, NeoBench contains pairs of sentences \((s_1,s_2)\) where \(s_2\) differs from \(s_1\) by the replacement of a target word with a neologism while ensuring same part of speech and same meaning of \(s_1\). 
Neologisms are collected between 2020-2023 and account for three types of vocabulary expansion, namely lexical, morphological, and semantic. 
Lexical neologisms include new words, phrases, and acronyms representing new concepts—e.g., `long covid'. 
Morphological neologisms instead are words that derive from existing words either through blending or splintering—e.g., `doomscrolling'.
Semantic neologisms refer to existing words with new meanings—e.g., `ice' to indicate petrol- or diesel-powered vehicles.
We manually annotate the Reddit sample of NeoBench as either hateful or non-hateful, reaching a substantial average inter-annotators' agreement (Cohen's Kappa \(=0.67\) \cite{cohen1960cohen_kappa_coefficient}) across three annotators.
We take the majority vote as groundtruth.
As a result, we have 341 annotated sentences \(s_1\) paired with their 341 counterfactuals \(s_2\) containing the neologisms in place of the target words.
We evaluate models using  \textbf{counterfactual invariance}, i.e., a formalization of the requirement that changing irrelevant parts of the input (i.e., replacing target words with neologisms) should not change model predictions \cite{veitch2021counterfactual_invariance}.
We decompose the counterfactual invariance into \emph{label flipping} (i.e., rate of how often the model flipped the label when seeing the counterfactual \(s_2\) wrt \(s_1\)) and \emph{hallucination} (i.e., rate of how often the model does not follow the instruction when given the counterfactual \(s_2\) but does follow the instruction when given \(s_1\)).
Mathematically, we define label flip =\(\frac{1}{N}\sum_{i=1}^{N} \mathbbm{1}( \hat{y}_i(s_1)\neq \hat{y}_i(s_2))\) and hallucination = \( \frac{1}{N}\sum_{i=1}^{N} \mathbbm{1} ( v(s_{2,i}) = 1 \land v(s_{1,i}) = 0 ) \) where \(v(\cdot)\) is \(1\) if model hallucinates, \(0\) otherwise. 
Ideally, we want language models to be robust against counterfactuals showing low label flip and hallucination rates, paired with high macro F1 score, which highlight their robustness to vocabulary changes and their ability to generalize to new words.

% \paragraph{Experiment 3: Reclaimed Language.}
% Hate speech can also evolve due to cultural shifts, e.g., words used offensively to target minorities can be reclaimed back by these communities. 
% To study this phenomenon, we use the Reclaimed Hate Speech dataset \cite{zsisku2024reclaimed_hate_speech} as it uniquely contains instances of popular slurs that have been reclaimed back in addition to instances with their standard offensive meaning.

% Looking at the F1 scores across labels, we can see how models are generally very good at detecting hateful instances but struggle a lot to capture reclaimed language, which is often misclassified as hateful.

\paragraph{Models.}
We zero-shot prompt 20 language models widely used in established hate speech and state-of-the-art research (Table \ref{tab:models_spec}). 
We use the verbalisation of \citet{plaza2023respectful-toxic}, which is shown to lead to the best performance in hate speech detection. 
% Models' details are in Appendix \ref{sec:appendix_models}. 
As baseline, we take the averaged scores of the latest versions of the TimeLMs collection fine-tuned for hate speech detection \cite{loureiro2022timelms,antypas2023supertweeteval}.
Cf. App. \ref{sec:appendix_ethical_nlp}.

\begin{table}[h]
\tiny
\centering
\begin{tabular}{l|ccc}
\hline
\textbf{Model} &  \textbf{\makecell[l]{Commercial}} & \textbf{\makecell[l]{Toxicity\\finetuned}} & \textbf{\makecell[l]{Data\\cutoff}} \\
\hline
FLAN-Alpaca & \redcross & \greencheck & -\\
% \hline
FLAN-T5 & \redcross & \greencheck & 2022/11\\ 
% \hline
mT0 & \redcross & \redcross & 2022/11 \\ 
% \hline
RoBERTa-dyna-r1 & \redcross & \greencheck & 2022/06\\ 
RoBERTa-dyna-r2 & \redcross & \greencheck & 2022/06\\
RoBERTa-dyna-r3 & \redcross & \greencheck & 2022/06\\
RoBERTa-dyna-r4 & \redcross & \greencheck & 2023/03\\
GPT-3.5-turbo & \greencheck & - & 2021/09\\
GPT-4o & \greencheck & - & 2023/10 \\
Moderation API & \greencheck & \greencheck & - \\
Perspective API & \greencheck & \greencheck & - \\
DeepSeek LLM & \redcross & - & - \\

\hline
\end{tabular}
\caption{Model overview. `-' if no available info.}
\label{tab:models_spec}
% \vspace{-5mm}
\end{table}

\paragraph{Static Benchmarks.}
We select established hate speech benchmarks: HateXplain \cite{mathew2021hatexplain}, Implicit Hate Corpus \cite{elsherief2021implicit_hate_corpus}, HateCheck \cite{rottger2021hatecheck}, and Dynabench \cite{kiela2021dynabench}.
Their selection is motivated by the fact that each static benchmark captures a distinct dimension of hate speech, thereby contributing to a more comprehensive assessment.
Specifically, we select the HateXplain and Implicit Hate Corpus datasets to account for the dimensions of, respectively, offensiveness and expressiveness of hate speech, as described in \citet{dibonaventura2025detection_explanation_LLM}. 
We include HateCheck because its construction aligns with the goals of Experiment 2, where models are tested on sentence pairs differing only in the target term. 
Similarly, HateCheck features sentences that differ only by the targeted group.
Finally, we select Dynabench as it is the only dynamic hate speech benchmark, built from adversarial examples collected across multiple rounds over time.
Note that the RoBERTa-dyna-r1/2/3/4 models \cite{vidgen2021roberta_hate_dynabench} in Table \ref{tab:models_spec} have been fine-tuned on four consecutive Dynabench rounds (i.e., dynamic adversarial training), which however increases the risk of creating unrealistic data distributions.
Table \ref{tab:dataset_spec} summarizes the datasets used in our time-sensitive and static evaluations. 

\begin{table}[h]
\tiny
\centering
\begin{tabular}{l|ccc}
\hline
\textbf{Dataset} &  \textbf{Size} & \textbf{Timestamp info} & \textbf{Timestamp period} \\
\hline
Singapore Online Attacks & \num{3000} & \greencheck & 2017-2022\\
NeoBench & \num{682} & \greencheck & 2020-2023\\
HateXplain & \num{1924} & \redcross & -\\
Implicit Hate Corpus & \num{2149} & \redcross & -\\
HateCheck & \num{3729} & \redcross & -\\
Dynabench & \num{4120} & \redcross & -\\
\hline
\end{tabular}
\caption{Dataset overview. `-' if not applicable.}
\label{tab:dataset_spec}
% \vspace{-5mm}
\end{table}

\begin{table*}[t]
\tiny
\centering
\begin{tabular}{|l||c|c|c|c|c|c||c|c|c|c|c|c||c|}
\hline
\textbf{Model} & \textbf{2017} & \textbf{2018} & \textbf{2019} & \textbf{2020} & \textbf{2021} & \textbf{2022} & \textbf{2017} & \textbf{2018} & \textbf{2019} & \textbf{2020} & \textbf{2021} & \textbf{2022} & \textbf{Mean} \\ \hline

FLAN-Alpaca-base & \colormap{0.1}{.1111} & \colormap{0.0}{.1026} & \colormap{0.99}{.1985} & \colormap{0.8}{.1789} & \colormap{0.53}{.1533} & \colormap{0.13}{.1148} & \colormap{0.0}{.7143} & \colormap{0.57}{.7853} & \colormap{0.96}{.8346} & \colormap{0.96}{.8347} & \colormap{0.99}{.8397} & \colormap{0.97}{.8364} & .5176 \\ \hline

FLAN-Alpaca-large & \colormap{0.99}{.6667} & \colormap{0.54}{.6023} & \colormap{0.33}{.5733} & \colormap{0.1}{.5383} & \colormap{0.45}{.5901} & \colormap{0.0}{.5265} & \colormap{0.0}{.7132} & \colormap{0.16}{.7222} & \colormap{0.63}{.7486} & \colormap{0.64}{.7491} & \colormap{0.97}{.7678} & \colormap{0.99}{.7694} & .6640 \\ \hline

FLAN-Alpaca-xl & \colormap{0.99}{.7258} & \colormap{0.49}{.6327} & \colormap{0.46}{.6268} & \colormap{0.29}{.5950} & \colormap{0.26}{.5896} & \colormap{0.0}{.5428} & \colormap{0.38}{.7069} & \colormap{0.0}{.6897} & \colormap{0.79}{.7254} & \colormap{0.99}{.7351} & \colormap{0.83}{.7274} & \colormap{0.62}{.7177} & .6679 \\ \hline

FLAN-T5-small & \colormap{0.0}{.0} & \colormap{0.0}{.0} & \colormap{0.0}{.0} & \colormap{0.0}{.0} & \colormap{0.0}{.0} & \colormap{0.0}{.0} & \colormap{0.0}{.6708} & \colormap{0.58}{.7625} & \colormap{0.83}{.8013} & \colormap{0.90}{.8118} & \colormap{0.95}{.8203} & \colormap{0.99}{.8278} & .3912 \\ \hline

FLAN-T5-base & \colormap{0.99}{.6557} & \colormap{0.50}{.5775} & \colormap{0.45}{.5698} & \colormap{0.33}{.5501} & \colormap{0.33}{.5513} & \colormap{0.0}{.4991} & \colormap{0.0}{.6441} & \colormap{0.38}{.6722} & \colormap{0.65}{.6917} & \colormap{0.99}{.7175} & \colormap{0.86}{.7069} & \colormap{0.62}{.6899} & .6272 \\ \hline

FLAN-T5-large & \colormap{0.99}{.7176} & \colormap{0.61}{.6332} & \colormap{0.43}{.5946} & \colormap{0.22}{.5472} & \colormap{0.31}{.5665} & \colormap{0.0}{.5000} & \colormap{0.99}{.6606} & \colormap{0.03}{.6540} & \colormap{0.78}{.6591} & \colormap{0.46}{.6569} & \colormap{0.14}{.6548} & \colormap{0.0}{.6538} & .6249 \\ \hline

FLAN-T5-xl & \colormap{0.99}{.7478} & \colormap{0.21}{.5969} & \colormap{0.47}{.6463} & \colormap{0.20}{.5961} & \colormap{0.18}{.5909} & \colormap{0.0}{.5571} & \colormap{0.94}{.7603} & \colormap{0.0}{.6723} & \colormap{0.99}{.7661} & \colormap{0.86}{.7530} & \colormap{0.80}{.7472} & \colormap{0.97}{.7631} & .6831 \\ \hline

mT0-small & \colormap{0.99}{.0435} & \colormap{0.0}{.0} & \colormap{0.41}{.0180} & \colormap{0.34}{.0147} & \colormap{0.23}{.0098} & \colormap{0.0}{.0} & \colormap{0.0}{.6716} & \colormap{0.64}{.7679} & \colormap{0.72}{.7798} & \colormap{0.97}{.8209} & \colormap{0.93}{.8123} & \colormap{0.99}{.8222} & .3967 \\ \hline

mT0-base & \colormap{0.0}{.0} & \colormap{0.67}{.0465} & \colormap{0.80}{.0559} & \colormap{0.99}{.0697} & \colormap{0.41}{.0289} & \colormap{0.52}{.0359} & \colormap{0.0}{.6588} & \colormap{0.60}{.7545} & \colormap{0.88}{.7994} & \colormap{0.97}{.8139} & \colormap{0.96}{.8123} & \colormap{0.99}{.8195} & .4079 \\ \hline

mT0-large & \colormap{0.99}{.5045} & \colormap{0.0}{.3669} & \colormap{0.63}{.4537} & \colormap{0.07}{.3769} & \colormap{0.06}{.3746} & \colormap{0.10}{.3811} & \colormap{0.0}{.5455} & \colormap{0.25}{.5737} & \colormap{0.99}{.6600} & \colormap{0.56}{.6094} & \colormap{0.82}{.6392} & \colormap{0.73}{.6290} & .5095 \\ \hline

mT0-xl & \colormap{0.0}{.2000} & \colormap{0.58}{.2718} & \colormap{0.99}{.3243} & \colormap{0.47}{.2581} & \colormap{0.67}{.2833} & \colormap{0.53}{.2657} & \colormap{0.0}{.6706} & \colormap{0.67}{.7692} & \colormap{0.92}{.8056} & \colormap{0.96}{.8115} & \colormap{0.98}{.8168} & \colormap{0.99}{.8177} & .5246 \\ \hline

RoBERTa-dyna-r1 & \colormap{0.95}{.4211} & \colormap{0.21}{.3519} & \colormap{0.99}{.4255} & \colormap{0.58}{.3864} & \colormap{0.84}{.4108} & \colormap{0.0}{.3322} & \colormap{0.0}{.7317} & \colormap{0.46}{.7813} & \colormap{0.92}{.8313} & \colormap{0.92}{.8313} & \colormap{0.99}{.8402} & \colormap{0.88}{.8277} & .5976 \\ \hline

RoBERTa-dyna-r2 & \colormap{0.72}{.3659} & \colormap{0.78}{.3692} & \colormap{0.32}{.3423} & \colormap{0.0}{.3236} & \colormap{0.69}{.3645} & \colormap{0.99}{.3824} & \colormap{0.0}{.6709} & \colormap{0.47}{.7248} & \colormap{0.78}{.7591} & \colormap{0.89}{.7716} & \colormap{0.99}{.7846} & \colormap{0.89}{.7726} & .5526 \\ \hline

RoBERTa-dyna-r3 & \colormap{0.41}{.3421} & \colormap{0.99}{.3571} & \colormap{0.0}{.3316} & \colormap{0.97}{.3569} & \colormap{0.10}{.3342} & \colormap{0.19}{.3364} & \colormap{0.0}{.6951} & \colormap{0.62}{.7722} & \colormap{0.82}{.7969} & \colormap{0.99}{.8188} & \colormap{0.97}{.8150} & \colormap{0.92}{.8093} & .5638 \\ \hline

RoBERTa-dyna-r4 & \colormap{0.99}{.5057} & \colormap{0.15}{.3859} & \colormap{0.18}{.3902} & \colormap{0.05}{.3724} & \colormap{0.08}{.3762} & \colormap{0.0}{.3652} & \colormap{0.0}{.7190} & \colormap{0.63}{.7771} & \colormap{0.88}{.7994} & \colormap{0.94}{.8051} & \colormap{0.99}{.8105} & \colormap{0.88}{.7996} & .5922 \\ \hline

GPT-3.5-turbo & \colormap{0.99}{.6846} & \colormap{0.55}{.6129} & \colormap{0.34}{.5799} & \colormap{0.15}{.5488} & \colormap{0.21}{.5590} & \colormap{0.0}{.5250} & \colormap{0.0}{.4598} & \colormap{0.09}{.4667} & \colormap{0.80}{.5233} & \colormap{0.47}{.4973} & \colormap{0.99}{.5389} & \colormap{0.33}{.4861} & .5402 \\ \hline

\textbf{GPT-4o} & \colormap{0.99}{.7619} & \colormap{0.69}{.7129} & \colormap{0.45}{.6742} & \colormap{0.23}{.6395} & \colormap{0.18}{.6311} & \colormap{0.0}{.6032} & \colormap{0.0}{.7368} & \colormap{0.30}{.7434} & \colormap{0.80}{.7542} & \colormap{0.99}{.7585} & \colormap{0.26}{.7424} & \colormap{0.22}{.7417} & \textbf{.7083} \\ \hline

Moderation API & \colormap{0.39}{.0645} & \colormap{0.0}{.0238} & \colormap{0.18}{.04120} & \colormap{0.99}{.1275} & \colormap{0.26}{.0507} & \colormap{0.38}{.0631} & \colormap{0.0}{.6742} & \colormap{0.57}{.7616} & \colormap{0.81}{.8000} & \colormap{0.98}{.8255} & \colormap{0.94}{.8203} & \colormap{0.99}{.8289} & .4235 \\ \hline

Perspective API & \colormap{0.90}{.4941} & \colormap{0.0}{.3486} & \colormap{0.75}{.4700} & \colormap{0.92}{.4966} & \colormap{0.99}{.5098} & \colormap{0.59}{.4431} & \colormap{0.0}{.7226} & \colormap{0.43}{.7774} & \colormap{0.86}{.8312} & \colormap{0.91}{.8080} & \colormap{0.99}{.8492} & \colormap{0.91}{.8374} & .6348 \\ \hline

DeepSeek LLM-7b & \colormap{0.99}{.7097} & \colormap{0.23}{.5000} & \colormap{0.36}{.5349} & \colormap{0.0}{.4356} & \colormap{0.06}{.4531} & \colormap{0.22}{.4957} & \colormap{0.08}{.1818} & \colormap{0.96}{.2667} & \colormap{0.59}{.2308} & \colormap{0.0}{.1739} & \colormap{0.99}{.2708} & \colormap{0.82}{.2532} & .3740 \\ \hline\hline

TimeLMs & .3620 & .3995 & .3505 & .3621 & .3080 & .3941 & .3547 & .3879 & .4104 & .4172 & .4128 & .4142 & .3722 \\ \hline

\end{tabular}
\caption{Time-sensitive Macro F1 for the hateful label (first block), non-hateful label (second block), and their macro-average (last column). Greener cells indicate higher scores; best score in \textbf{bold}. Std deviations in App. \ref{sec:appendix_std_dev}.}
\label{tab:time-sensitive-f1}
\end{table*}

\section{Findings}

\paragraph{Language models exhibit short- and long-term volatility in hate speech detection across years.}

Table \ref{tab:time-sensitive-f1} presents time-sensitive macro F1 by label, and their average in the last column. 
Although all models have data cutoffs equal to or later than 2021, they fail to generalise well to time-sensitive shifts occurring between 2017 and 2022 as shown by the significant changes in the macro F1 scores year by year for both labels. 
%For instance, mT0-large has macro F1 for the hateful label equal to .5045 in 2017, which drops to .3669 the following year, goes back to .4537 in 2019, and drops again to .3769 in 2020.
In addition to this volatile pattern year-by-year, we observe a long-term pattern: most language models exhibit a decreasing performance in detecting hateful instances and an increasing performance in detecting non-hateful content between 2017 and 2022.
For example, mT0-large has macro F1 equal to .5045 and .5455 for hateful and non-hateful labels, respectively, in 2017.
By 2022, it has instead .3811 and .6290. 
As hate speech classifiers suffer from lexical overfit (e.g., \citet{attanasio2022entropy_bias_hate}), we argue they tend to over-rely on older lexical associations for which there is more evidence in the data (e.g., `gammon' as ham), and thus fail to recognise newer/emerging associations (e.g., `gammon' as insult).
%For example, FLAN-T5-base reaches .6557 macro F1 in 2017, which decreases over the years, reaching .4991 in 2022. 
%On the other hand, most language models become better at detecting non-hateful content between 2017 and 2022; e.g., FLAN-Alpaca-base has macro F1 equal to .7143 in 2017 which increases over the years, reaching .8364 in 2022. 
Clearly, this short-term and long-term volatility of language models in evolving hate speech detection poses real concerns regarding the safety robustness of these models. 
Interestingly, dynamic adversarial training does not make models more robust to time-sensitive shifts: RoBERTa-dyna-r2/3/4 models which have been fine-tuned on more adversarial examples than RoBERTa-dyna-r1 have lower time-sensitive macro F1 than the latter. 
This corroborates previous research showing that training on adversarially-collected data for QA tasks was detrimental to performance on non-adversarially collected data \cite{bartolo2020beat_ai}.
%, and empirically shows that adversarial training increases the risk of creating unrealistic data distributions which do not generalise over time. 
For the other non-adversarially trained models instead, model size improves the overall time-sensitive macro F1 score. 
The time-sensitive baseline is more robust across years and labels but overall performs similarly to small LLMs and DeepSeek LLM.
GPT-4o reaches the highest time-sensitive performance.

\paragraph{Language models are sensitive to counterfactuals containing neologisms.}

Table \ref{tab:counterfactual_invariance} shows how often models flip the predicted label and generate hallucinations when they see the counterfactual with respect to the reference sentence, and the macro F1 performance in detecting hate speech in those sentences. 
The label flip rates are surprisingly high, considering that models' cutoffs have some overlap with the timeframe from which the neologisms were sampled: 6 out of 20 models flip the label more than 10\% of the time.\footnote{We also controlled for time to measure the potential impact of data contamination, and found no evidence (cf. Table \ref{tab:label_flip_by_year} and Table \ref{tab:hallucination_by_year} in App. \ref{sec:appendix_more_hallucination}).} 
Interestingly, counterfactuals have a greater impact on making the model change its predicted label than on generating a non-response, as evidenced by the lower hallucination rates compared to label flips.
Moreover, model size lowers the tendency to hallucinate but does not necessarily improve the label flip rate.
For instance, FLAN-Alpaca-xl has 0\% hallucination vs. 10.88\% of FLAN-Alpaca-large but flips the label more frequently (14.14\% vs. 3.98\%). 
Similarly, GPT-4o has a worse label flip rate than smaller and/or earlier models like RoBERTa-dyna-r2/3/4. 
One reason for this behaviour may be excessive memorization, which is more likely
to occur with larger model sizes \cite{kiyomaru2024LLMs_memorization,tirumala2022LLMs_memorization,carlini2022LLMs_memorization}.
Consistently with the findings of Experiment 1, RoBERTa-dyna-r2/3/4 are less robust to counterfactuals than RoBERTa-dyna-r1, which has lower label flip rate and higher macro F1 score.
Additionally, the TimeLMs baseline is more robust to language evolution, even though most LLMs outperform it in classification performance.
With the exception of DeepSeek LLM (which, however, has high hallucination rates; cf. Table \ref{tab:hall_rates_details}), a label flip rate of 0 occurs when a model outputs the same label for all texts; so if we exclude these models, the best one is Perspective API with a minimal label flip rate and the highest macro F1. 
Moreover, we investigate label flip and hallucination rates by type of vocabulary expansion in Table \ref{tab:label_flip_by_vocabulary_type} and Table \ref{tab:hallucination_by_vocabulary_type}, respectively. 
We found that on average models flip the label more often if the counterfactual sentence contains a morphological neologism whereas they tend to hallucinate more often in case of lexical neologism.

\begin{table}[h!]
\tiny
\centering
\begin{tabular}{|l|c|c|c|c|}
\hline
\textbf{Model} & \textbf{Label Flip (\%)} & \textbf{Hallucination (\%)} & \textbf{Macro F1} \\ \hline
FLAN-Alpaca-base & 0.65 & 3.82 & .5189 \\ \hline
FLAN-Alpaca-large & 3.98 & 10.88 & .5626 \\ \hline
FLAN-Alpaca-xl & 14.14 & 0.00 & .5344 \\ \hline
FLAN-T5-small & 0.00 & 2.06 & .4851 \\ \hline
FLAN-T5-base & 11.24 & 0.00 & .4774 \\ \hline
FLAN-T5-large & 15.96 & 0.88 & .4742 \\ \hline
FLAN-T5-xl & 13.99 & 0.88 & .6002 \\ \hline
mT0-small & 0.00 & 4.41 & .4881 \\ \hline
mT0-base & 0.59 & 0.00 & .4824 \\ \hline
mT0-large & 14.12 & 0.00 & .3383 \\ \hline
mT0-xl & 3.53 & 0.00 & .5261 \\ \hline
RoBERTa-dyna-r1 & 3.53 & - & .6451 \\ \hline
RoBERTa-dyna-r2 & 5.88 & - & .5931 \\ \hline
RoBERTa-dyna-r3 & 5.00 & - & .5437 \\ \hline
RoBERTa-dyna-r4 & 6.47 & - & .5737 \\ \hline
GPT-3.5-turbo & 14.93 & 0.88 & .4885 \\ \hline
GPT-4o & 9.44 & 0.00 & .6636 \\ \hline
Moderation API & 0.00 & - & .4841 \\ \hline
\textbf{Perspective API} & \textbf{2.94} & - & \textbf{.7067} \\ \hline
DeepSeek LLM-7b & 0.00 & 1.17 & .2500 \\ \hline\hline
TimeLMs & 0.30 & - & .2929 \\ \hline
\end{tabular}
\caption{Label Flip and Hallucination rates, and Macro F1. Best score in \textbf{bold}. `-' if not applicable.}
\label{tab:counterfactual_invariance}
\end{table}

\paragraph{High scores in static evaluations do not necessarily translate to time-sensitive evaluations.}

Table \ref{tab:corr_rankings} shows the Spearman's rank correlation coefficient of models' ranking between static and time-sensitive evaluations, paired with their confidence intervals.
These coefficients are computed by comparing the rankings of the best performing models between each possible pair of static and time-sensitive evaluations. 
We use the rankings on the four benchmarks in Table \ref{tab:f1_hatecheck}-\ref{tab:f1_implicit_hate} in App. \ref{sec:appendix_benchmarks} for the static evaluations whereas we use those in Table \ref{tab:time-sensitive-f1} and Table \ref{tab:counterfactual_invariance} for the time-sensitive evaluations.
The confidence intervals are computed setting \(\alpha=0.10\), which means that there is a 90\% confidence that the intervals contain the true population correlation coefficients between static and time-sensitive evaluations.
There is a clear misalignment between the two types of evaluations. 
Overall, there is a negative correlation between static evaluations and Experiment 1, indicating that models that perform the best in static benchmarks are not the most robust to time-sensitive shifts. 
Similarly, high scores in static evaluations do not necessarily imply high scores in Experiment 2, as correlation is on average negative or close to zero.
On the other hand, static hate speech benchmarks show a positive, non-negligible correlation among each other, with an average correlation coefficient equal to 0.36 (cf. Table \ref{tab:corr_static_benchs} and App. \ref{sec:appendix_benchmarks}).
In other words, while performance on a static hate speech benchmark is aligned to the performance on another static benchmark, the same does not hold for time-sensitive evaluations. 
Evolving hate speech introduces variability that static benchmarks fail to capture, making them an unreliable predictor over time.

% This is the key for the title, showing that high scores on static benchmarks can be misleading, and we need more structured research on evaluating language models over time! Which brings us to develop the benchmark for abusive language over time, grounded on time-sensitive linguistic graph.

% \begin{table}[h!]
% \tiny
% \centering
% \begin{tabular}{|l|c|c|}
% \hline
% \textbf{$\downarrow$ Static / Time-sensitive $\rightarrow$} & \textbf{Experiment 1} & \textbf{Experiment 2} \\ \hline
% \textbf{HateCheck} & -0.0070 & -0.0707 \\ \hline
% \textbf{Dynabench} & -0.1456 & -0.3053 \\ \hline
% \textbf{HateXplain} & -0.2368 & -0.1865 \\ \hline
% \textbf{Implicit Hate} & -0.3649 & 0.1909 \\ \hline
% \end{tabular}
% \caption{Spearman coefficients between static and time-sensitive evaluations. Confidence intervals in App. \ref{sec:appendix_confidence}.}
% \label{tab:corr_rankings}
% \end{table}

\begin{table}[h!]
\scriptsize
\centering
\begin{tabular}{cc|cc}
\multirow{12}{*}{\rotatebox{90}{\textbf{Static}}} & & \multicolumn{2}{c}{\textbf{Time-sensitive}} \\
 &  & \textbf{Experiment 1} & \textbf{Experiment 2} \\
\hline
 & \textbf{HateCheck} & \makecell{-0.2662 \\ {\tiny \textit{(-0.586, 0.126)}}} & \makecell{-0.0707 \\ {\tiny \textit{(-0.438, 0.317)}}} \\ \cline{3-4}
 
 & \textbf{Dynabench} & \makecell{-0.1549 \\ {\tiny \textit{(-0.504, 0.238)}}} & \makecell{-0.3053 \\ {\tiny \textit{(-0.613, 0.083)}}} \\ \cline{3-4} 
 
 & \textbf{HateXplain} & \makecell{-0.2541 \\ {\tiny \textit{(-0.578, 0.138)}}} & \makecell{-0.1865 \\ {\tiny \textit{(-0.528, 0.207)}}} \\ \cline{3-4} 
 
 & \textbf{Implicit Hate} & \makecell{-0.2812 \\ {\tiny \textit{(-0.597, 0.110)}}} & \makecell{0.1909 \\ {\tiny \textit{(-0.203, 0.532)}}} \\ \cline{1-4} 
 
\end{tabular}
\caption{Spearman coefficients between static and time-sensitive evaluations. 90\% confidence intervals shown below each value. Cf. App. \ref{sec:appendix_confidence}.}
\label{tab:corr_rankings}
\end{table}

\section{Related Work}
\paragraph{Language evolution and model training.}
The evolving nature of language has attracted a great interest in the NLP community to address the so-called temporal bias, i.e., decreasing performance over time \cite{alkhalifa2023building_for_tomorrow}, by training models to adapt to newer data \cite{dhingra2022time_aware_lms, lazaridou2021mind_the_gap, rottger2021temporal_adaptation, jang2021towards_continual_learning}, historical data \cite{qiu2022histbert, martinc2020historical_contextual_embeddings}, or to be constrained to a specific time period \cite{drinkall2024time_machine_gpt}.  
In the hate speech domain, this has led to the proposal of several approaches to train time-sensitive hate speech classifiers 
% (\citet{qian2021lifelong_learning_hate},\citet{mcgillivray2022time_sensitive_offensive_detection}, \citet{florio2020time_hate}, \citet{jin2023examining_temporal_bias}, \textit{inter alia}).
like lifelong learning \cite{qian2021lifelong_learning_hate}, time-sensitive knowledge-injection \cite{mcgillivray2022time_sensitive_offensive_detection}, random vs. chronological data splits \cite{florio2020time_hate}, temporal adaptation \cite{jin2023examining_temporal_bias}. 
These studies focus either on BERT-based models or non-neural ones. 
Instead, we investigate the temporal bias of 20 state-of-the-art LLMs in hate speech detection in two scenarios of language evolution.

\paragraph{Language evolution and model benchmarking.}
While the implications of evolving hate speech in model training have been widely investigated, its implications in model benchmarking have been overlooked. 
This gap is especially important given the rise of LLMs, where hate speech benchmarks are often embedded in safety evaluations \cite{ying2024safebench}. 
Remarkably, we provide empirical evidence of the unreliability of static hate speech benchmarks over time due to evolving hate speech, thus calling for time-sensitive linguistic benchmarks in this domain.
This type of linguistic benchmarks is scarce as most studies focus on encyclopedic and commonsense knowledge to evaluate models' ability to understand factual changes regarding entities and events (e.g., \citet{fatemi2024google_test_of_time, wang2024tram, tan2023tempreason}) rather than language changes.
A loosely related study is \citet{pozzobon2023perspective_api_over_time} showing that Perspective API yields unreliable toxicity predictions over time due to model updates.
Instead, we measure the implications due to evolving language.

\section{Conclusions}
This study is the first to investigate the impact of evolving language on hate speech benchmarking. 
We design two time-sensitive experiments and metrics to evaluate 20 language models widely adopted in state-of-the-art research. 
We found that language models are not robust to evolving hate speech as they exhibit short- and long-term volatility to time-sensitive shifts in Experiment 1 and sensitivity to counterfactuals containing neologisms in Experiment 2. 
% Even GPT-4o, which is the most recent language model with training data until October 2023, has a label flip rate of 9.44\% when exposed to counterfactuals with neologisms coined between 2020 and 2023.
Interestingly, dynamic adversarial training does not help models generalise in evolving scenarios. 
% beyond their distributional knowledge. 
% Although it helps make models more robust to adversarial attacks, it risks to generate unrealistic data distributions which do not reflect the actual distribution of language over time.
Finally, we provide empirical evidence of the misalignment between static and time-sensitive evaluations, as we found negative or close to zero correlations between the two, which opens up important concerns about the reliability of current hate speech benchmarks in the future.

In light of our findings, we advocate for time-sensitive linguistic benchmarks to reliably evaluate models' safety in the hate speech domain. 
Examples might include our proposed time-sensitive metrics or more structured approaches similar to those recently developed for evolving encyclopedic knowledge (e.g., Test-of-Time \cite{fatemi2024google_test_of_time}). Future techniques could explore continual learning to enable LLMs to adapt to evolving hate speech, %meta-learning to help models generalize from past language shifts and address new hate speech patterns, 
and context-aware detection to capture subtle shifts in meaning driven by cultural or political events.

% Since static hate speech benchmarks do not account for evolving language, they return only a partial view of model safety and can be misleading to evaluate model safety over time as there is not a strong positive correlation with time-sensitive evaluations.

\section{Limitations}
We are aware of the following limitations. 
\textbf{(1)} We recognize hate speech as a multilingual problem. 
However, in this paper we prioritized English because resources for English hate speech are easily available and well-developed, providing a strong foundation for our study. 
Extending to multilingualism is an interesting direction for future work. 
\textbf{(2)} Although we chose established, well-documented and public datasets for our analyses, hate speech datasets inherently contain bias and noise due to the subjective nature of annotations and the social context in which the data were collected.
\textbf{(3)} 
We consider two aspects of language evolution, namely time-sensitive shifts and vocabulary expansion. 
We did not disentangle the individual contributions of sub-categories of time-sensitive shifts, such as polarity or topical, since they are notoriously hard to isolate and out of scope for this paper. 
However, it is an interesting direction for future work. 
\textbf{(4)} Continuous data collection of social media content is a challenge in current research based on social media platforms. 
This difficulty challenges performing Experiment 1 over time in the future, but it does not impact the ability of carrying out Experiment 2, which instead can be done using established linguistic resources like Oxford English Dictionary, Wiktionary, Urban Dictionary.

\section*{Acknowledgements}

This work was supported by the UK Research and Innovation [grant number EP/S023356/1] in the UKRI Centre for Doctoral Training in Safe and Trusted Artificial Intelligence (www.safeandtrustedai.org). 
Moreover, CDB work was supported by The Alan Turing Institute’s Enrichment Scheme.

\section*{Author Contribution Statement}

Authors contributed to the project as follows. 
\textbf{Project Conception}: Di Bonaventura, McGillivray, Meroño-Peñuela. 
\textbf{Literature Review}: Di Bonaventura. 
\textbf{Experimental Design}: Di Bonaventura. 
\textbf{Analysis Advisory}: McGillivray, He. 
\textbf{Manual Annotation}: Di Bonaventura, McGillivray, Meroño-Peñuela. 
\textbf{Results and Codebase}: Di Bonaventura. 
\textbf{Manuscript Writing}: Di Bonaventura. 
\textbf{Manuscript Editing and Feedback}: Everyone.

\bibliography{custom}

\appendix

\setcounter{table}{0}
\renewcommand{\thetable}{A\arabic{table}}
\setcounter{figure}{0}
\renewcommand{\thefigure}{A\arabic{figure}}

\section{Ethical NLP Research}
\label{sec:appendix_ethical_nlp}

\paragraph{Data.} 
We use publicly available datasets for our experiments, which ensure anonymized content. 
The use of these datasets is consistent with their terms for use and intended use. 
They only cover English. 
For Experiment 1 and 2, the size of the data used were \num{3000} and \num{682}, respectively.
The size of the static hate speech datasets are: \num{3729} (HateCheck), \num{1924} (HateXplain), \num{4120} (Dynabench), and \num{2149} (Implicit Hate).
We use the test sets.

\paragraph{Models. }
For our experiments, we choose widely used language models for hate speech research, considering a variety of characteristics like open-source vs. commercial models, encoder-decoder vs. decoder-only models, previously toxicity fine-tuned vs. not previously toxicity fine-tuned, and with different training data cutoff dates. 
Next, we briefly describe each model we analysed: 
\begin{itemize}
    \item FLAN-Alpaca \cite{bhardwaj2023flan_alpaca}: an instruction-tuned derivative of FLAN-T5, further instruction fine-tuned on Alpaca \cite{alpaca} dataset. It was previously finetuned for toxicity detection.
    \item FLAN-T5 \cite{wei2021flan_dataset}: an instruction fine-tuned derivative of T5 \cite{xue2021mt5} using the dataset FLAN \cite{wei2021flan_dataset}. It was previously toxicity finetuned. 
    \item mT0 \cite{muennighoff-etal-2023-mt0}: an instruction fine-tuned derivative of mT5 \cite{xue2021mt5} finetuned on xP3 dataset \cite{muennighoff-etal-2023-mt0}.
    \item RoBERTa-dyna-r1/2/3/4 \cite{vidgen2021roberta_hate_dynabench}: iterative versions of RoBERTa \cite{liu2019roberta} fine-tuned dynamically on increasingly refined training data from Dynabench \cite{kiela2021dynabench}.
    \item GPT-3.5-turbo\footnote{\url{https://platform.openai.com/docs/models/gp\#gpt-3-5-turbo}}: cost-efficient, highly optimized version of OpenAI’s GPT-3.5.
    \item GPT-4o\footnote{\url{https://platform.openai.com/docs/models\#gpt-4o}}: specialized variant of OpenAI's GPT-4 designed for optimized performance. 
    \item Moderation API\footnote{\url{https://platform.openai.com/docs/guides/moderation/overview}}: the OpenAI tool to identify potentially harmful content in text and images. We use the text-moderation-latest as underlying model. 
    \item Perspective API\footnote{\url{https://perspectiveapi.com/}}: a machine learning-based API by Google that scores text for perceived toxicity.
    \item DeepSeek LLM\footnote{\url{https://huggingface.co/deepseek-ai/deepseek-llm-7b-base}}: 7B language model with multi-head attention trained on 2 trillion tokens from scratch in English and Chinese, developed by DeepSeek.
    \item TimeLMs \cite{loureiro2022timelms,antypas2023supertweeteval}: a time-sensitive language model, based on the RoBERTa architecture, for NLP research in the  social media domain. It is continuously trained on social media over regular intervals for researching language model degradation, as well as cultural shifts affecting language usage on social media. We use the versions finetuned for hate speech detection as baseline. 
\end{itemize}

\paragraph{Experimental Setup.}
In all our analyses, we use zero-shot prompting. 
For each model, we use the default set of parameters; we refer the reader to the original papers for more details. 
The prompts are designed following the instructions used in  \citet{plaza2023respectful-toxic}, which are shown to be the most suitable for hate speech detection. 
Specifically, the instruction passed to the models is \textit{``Classify the input text as `respectful' or `toxic'.''}. 
Then, for the computations of the macro F1 we disregard outputs that were not as expected, i.e., that did not follow the instruction of answering with one word either `toxic' or `respectful'. 
The xl sizes of the models were loaded using 8bit quantization. 
We will release the code upon acceptance of the paper.

\paragraph{Manual Annotation.}
Three authors of this paper were involved in the manual annotation of the Reddit sample of NeoBench. 
Annotators are AI researchers, familiar with the domain of hate speech, and with English language.
They were presented sentences and asked to annotate whether the sentence was hateful or non-hateful. 
We take the majority vote as groundtruth.

\section{Experiment 1}
\label{sec:appendix_std_dev}
Following, we report additional results for Experiment 1. Specifically, Table \ref{tab:std_deviation_experiment1} shows the standard deviation of macro F1 for the hateful and non-hateful label over time.

\begin{table}[h!]
\tiny
\centering
\begin{tabular}{|l|c|c|}
\hline
\textbf{Model} & \textbf{Std dev `hateful'} & \textbf{Std dev `non-hateful'} \\ \hline
FLAN-Alpaca-base & 0.0363 & 0.0457 \\ \hline
FLAN-Alpaca-large & 0.0460 & 0.0211 \\ \hline
FLAN-Alpaca-xl & 0.0561 & 0.0150 \\ \hline
FLAN-T5-small & 0.00 & 0.0541 \\ \hline
FLAN-T5-base & 0.0468 & 0.0239 \\ \hline
FLAN-T5-large & 0.0690 & 0.0026 \\ \hline
FLAN-T5-xl & 0.0618 & 0.0325 \\ \hline
mT0-small & 0.0147 & 0.0522 \\ \hline
mT0-base & 0.0220 & 0.0568 \\ \hline
mT0-large & 0.0514 & 0.0392 \\ \hline
mT0-xl & 0.0368 & 0.0524  \\ \hline
RoBERTa-dyna-r1 & 0.0352 & 0.0388  \\ \hline
RoBERTa-dyna-r2 & 0.0194 & 0.0389 \\ \hline
RoBERTa-dyna-r3 & 0.0104 & 0.0429 \\ \hline
RoBERTa-dyna-r4 & 0.0483 & 0.0314  \\ \hline
GPT-3.5-turbo & 0.0522 & 0.0285  \\ \hline
GPT-4o & 0.0536 & 0.0076  \\ \hline
Moderation API & 0.0323 & 0.0546  \\ \hline
Perspective API & 0.0544 & 0.0451 \\ \hline
DeepSeek LLM-7b & 0.0902 & 0.0388 \\ \hline\hline
TimeLMs & 0.0302 & 0.0222 \\ \hline
\end{tabular}
\caption{Standard deviation of macro F1 for hateful and non-hateful label over time.}
\label{tab:std_deviation_experiment1}
\end{table}

\section{Experiment 2}
\label{sec:appendix_more_hallucination}
Following, we report additional results for Experiment 2. 

In Table \ref{tab:label_flip_by_year} and Table \ref{tab:hallucination_by_year}, we measure the same metrics of Table \ref{tab:counterfactual_invariance} while controlling for time. 
Since the NeoBench dataset provides timestamps for each pair \((s_1, s_2)\) marking the emergence of the neologism, we verified that label flip and hallucination rates remain comparable across years. 
This helps address concerns about potential data contamination, which would likely have resulted in a peak of these metrics in later years due to the partial overlap between the neologisms’ timeframe and the models' training cutoff dates. 
Our analysis found no evidence of such contamination, as the metrics remain overall stable across different years. 
Nevertheless, data contamination remains a general challenge in NLP research, and it is difficult to rule out entirely due to the lack of transparency regarding most models' training data.
Results are shown in Table \ref{tab:label_flip_by_year} and Table \ref{tab:hallucination_by_year} for label flip and hallucination rates, respectively. 
For this computation, we ruled out pairs whose timestamp information was missing in NeoBench.

\begin{table}[h!]
\tiny
\centering
\begin{tabular}{|l|c|c|c|c|c|}
\hline
\textbf{Model} & \textbf{2020} & \textbf{2021} & \textbf{2022} & \textbf{2023} \\ \hline
FLAN-Alpaca-base & 0.00 & 0.00 & 2.50 & 0.00 \\ \hline
FLAN-Alpaca-large & 6.58 & 0.00 & 4.92 & 0.00 \\ \hline
FLAN-Alpaca-xl & 13.21 & 15.56 & 16.67 & 0.00 \\ \hline
FLAN-T5-small & 0.00 & 0.00 & 0.00 & 0.00 \\ \hline
FLAN-T5-base & 12.27 & 8.89 & 14.45 & 0.00 \\ \hline
FLAN-T5-large & 11.43 & 19.32 & 20.96 & 0.00 \\ \hline
FLAN-T5-xl & 13.33 & 16.09 & 13.33 & 12.50 \\ \hline
mT0-small & 0.00 & 0.00 & 0.00 & 0.00 \\ \hline
mT0-base & 1.89 & 0.00 & 0.00 & 0.00 \\ \hline
mT0-large & 18.89 & 11.11 & 13.33 & 12.50 \\ \hline
mT0-xl & 6.60 & 2.22 & 3.33 & 0.00 \\ \hline
RoBERTa-dyna-r1 & 2.83 & 3.33 & 2.22 & 12.50 \\ \hline
RoBERTa-dyna-r2 & 7.55 & 7.78 & 4.44 & 0.00 \\ \hline
RoBERTa-dyna-r3 & 6.60 & 5.56 & 2.22 & 0.00 \\ \hline
RoBERTa-dyna-r4 & 9.43 & 5.56 & 3.33 & 0.00 \\ \hline
GPT-3.5-turbo & 12.38 & 21.84 & 12.36 & 12.50 \\ \hline
GPT-4o & 10.38 & 7.78 & 8.99 & 0.00 \\ \hline
Moderation API & 0.00 & 0.00 & 0.00 & 0.00 \\ \hline
Perspective API & 1.89 & 3.33 & 3.33 & 12.50 \\ \hline
DeepSeek LLM-7b & - & 0.00 & - & 0.00 \\ \hline
\end{tabular}
\caption{Label Flip Rates (in \%) by year. `-' if not applicable as the model did not generate any outputs as expected.}
\label{tab:label_flip_by_year}
\end{table}

\begin{table}[h!]
\tiny
\centering
\begin{tabular}{|l|c|c|c|c|c|}
\hline
\textbf{Model} & \textbf{2020} & \textbf{2021} & \textbf{2022} & \textbf{2023} \\ \hline
FLAN-Alpaca-base & 4.72 & 2.22 & 4.44 & 0.00 \\ \hline
FLAN-Alpaca-large & 7.55 & 14.44 & 10.00 & 25.00 \\ \hline
FLAN-Alpaca-xl & 0.00 & 0.00 & 0.00 & 0.00 \\ \hline
FLAN-T5-small & 0.94 & 5.56 & 1.11 & 0.00 \\ \hline
FLAN-T5-base & 0.00 & 0.00 & 0.00 & 0.00 \\ \hline
FLAN-T5-large & 0.00 & 0.00 & 2.22 & 0.00 \\ \hline
FLAN-T5-xl & 0.94 & 2.22 & 0.00 & 0.00 \\ \hline
mT0-small & 9.43 & 1.11 & 1.11 & 0.00 \\ \hline
mT0-base & 0.00 & 0.00 & 0.00 & 0.00 \\ \hline
mT0-large & 0.00 & 0.00 & 0.00 & 0.00 \\ \hline
mT0-xl & 0.00 & 0.00 & 0.00 & 0.00 \\ \hline
RoBERTa-dyna-r1 & - & - & - & - \\ \hline
RoBERTa-dyna-r2 & - & - & - & - \\ \hline
RoBERTa-dyna-r3 & - & - & - & - \\ \hline
RoBERTa-dyna-r4 & - & - & - & - \\ \hline
GPT-3.5-turbo & 0.00 & 2.22 & 1.11 & 0.00 \\ \hline
GPT-4o & 0.00 & 0.00 & 0.00 & 0.00 \\ \hline
Moderation API & - & - & - & - \\ \hline
Perspective API & - & - & - & - \\ \hline
DeepSeek LLM-7b & 0.94 & 0.00 & 2.22 & 0.00 \\ \hline
\end{tabular}
\caption{Hallucination Rates (in \%) by year. `-' if not applicable as models are non-generative.}
\label{tab:hallucination_by_year}
\end{table}

Moreover, we compute label flip and hallucination rates in Experiment 2 by type of vocabulary expansion. 
Specifically, Table \ref{tab:label_flip_by_vocabulary_type} contains label flip rates whereas Table \ref{tab:hallucination_by_vocabulary_type} contains hallucination rates.
From one hand, models on average flip the label more often if the counterfactual sentence contains a morphological vocabulary expansion (average label flip rate equal to 6.54\%) rather than lexical (6.40\%) or semantic ones (5.34\%).
On the other hand, models tend to hallucinate more often in cases of lexical vocabulary expansion (average hallucination rate equal to 2.12\%) rather than morphological (1.58\%) and semantic ones (1.82\%).

\begin{table}[h!]
\tiny
\centering
\begin{tabular}{|l|c|c|c|c|}
\hline
\textbf{Model} & \textbf{Lexical} & \textbf{Morphological} & \textbf{Semantic} \\ \hline
FLAN-Alpaca-base & 2.06 & 0.00 & 0.00 \\ \hline
FLAN-Alpaca-large & 0.00 & 6.20 & 4.17 \\ \hline
FLAN-Alpaca-xl & 14.81 & 15.68 & 8.51 \\ \hline
FLAN-T5-small & 0.00 & 0.00 & 0.00 \\ \hline
FLAN-T5-base & 10.38 & 10.81 & 14.89\\ \hline
FLAN-T5-large & 15.24 & 18.68 & 6.67 \\ \hline
FLAN-T5-xl & 22.22 & 11.00 & 6.52\\ \hline
mT0-small & 0.00 & 0.00 & 0.00 \\ \hline
mT0-base & 0.00 & 0.54 & 2.13\\ \hline
mT0-large & 13.89 & 14.59 & 12.77\\ \hline
mT0-xl & 3.70 & 3.24 & 4.26 \\ \hline
RoBERTa-dyna-r1 & 5.56 & 2.70 & 2.13 \\ \hline
RoBERTa-dyna-r2 & 7.41 & 4.86 & 6.38 \\ \hline
RoBERTa-dyna-r3 & 5.56 & 4.86 & 4.26 \\ \hline
RoBERTa-dyna-r4 & 6.48 & 5.95 & 8.51 \\ \hline
GPT-3.5-turbo & 12.38 & 16.94 & 12.77\\ \hline
GPT-4o & 6.48 & 11.41 & 8.51 \\ \hline
Moderation API & 0.00 & 0.00 & 0.00 \\ \hline
Perspective API & 1.85 & 3.24 & 4.26\\ \hline
DeepSeek LLM-7b & 0.00 & 0.00 & 0.00 \\ \hline
\end{tabular}
\caption{Label Flip Rates (in \%) by type of vocabulary expansion. `-' if not applicable as the model did not generate any outputs as expected.}
\label{tab:label_flip_by_vocabulary_type}
\end{table}

\begin{table}[h!]
\tiny
\centering
\begin{tabular}{|l|c|c|c|c|}
\hline
\textbf{Model} & \textbf{Lexical} & \textbf{Morphological} & \textbf{Semantic} \\ \hline
FLAN-Alpaca-base & 4.63 & 2.70 & 6.38 \\ \hline
FLAN-Alpaca-large & 10.19 & 11.89 & 8.51 \\ \hline
FLAN-Alpaca-xl & 0.00 & 0.00 & 0.00\\ \hline
FLAN-T5-small & 2.78 & 1.08 & 4.26\\ \hline
FLAN-T5-base & 0.00 & 0.00 & 0.00\\ \hline
FLAN-T5-large & 1.85 & 0.00 & 2.13\\ \hline
FLAN-T5-xl & 0.00 & 1.08 & 2.13 \\ \hline
mT0-small & 5.56 & 4.32 & 2.13 \\ \hline
mT0-base & 0.00 & 0.00 & 0.00\\ \hline
mT0-large & 0.00 & 0.00 & 0.00\\ \hline
mT0-xl & 0.00 & 0.00 & 0.00 \\ \hline
RoBERTa-dyna-r1 & - & - & - \\ \hline
RoBERTa-dyna-r2 & - & - & - \\ \hline
RoBERTa-dyna-r3 & - & - & - \\ \hline
RoBERTa-dyna-r4 & - & - & - \\ \hline
GPT-3.5-turbo & 1.85 & 0.54 & 0.00\\ \hline
GPT-4o & 0.00 & 0.00 & 0.00 \\ \hline
Moderation API & - & - & - \\ \hline
Perspective API & - & - & - \\ \hline
DeepSeek LLM-7b & 2.78 & 0.54 & 0.00 \\ \hline
\end{tabular}
\caption{Hallucination Rates (in \%) by type of vocabulary expansion. `-' if not applicable as the model are non-generative.}
\label{tab:hallucination_by_vocabulary_type}
\end{table}

In addition to the hallucination rates shown in Table \ref{tab:counterfactual_invariance}, we compute hallucination rates considering reference and counterfactual sentences, and only counterfactual sentences. 
Mathematically, we define the former as \( hal_{s_1,s_2} = \frac{1}{N}\sum_{i=1}^{N} \mathbbm{1} ( v(s_{2,i}) = 1 \lor v(s_{1,i}) = 1 ) \) and the latter as \( hal_{s_2} = \frac{1}{N}\sum_{i=1}^{N} \mathbbm{1} ( v(s_{2,i}) = 1 ) \). 
We consider hallucination any answer given by the model which does not follow the instruction given in the prompt—e.g., when the model repeats the instruction without providing any answer regarding the classification. 
Results are shown in Table \ref{tab:hall_rates_details}.  
Overall, hallucination rates are surprisingly high: 5 out of 14 models\footnote{Non-generative models are disregarded in this computation.} hallucinate more than 10\% of the time on either reference or counterfactual sentences as shown in the first column.
This hallucination is mostly driven by the presence of counterfactual sentences, as shown in the last column.
In particular, DeepSeek LLM shows incredibly high hallucination rates compared to the other language models. 

\begin{table}[h!]
\tiny
\centering
\begin{tabular}{|l|c|c|c|}
\hline
\textbf{Model} & \textbf{Hal\textsubscript{s1,s2} (\%)} & \textbf{Hal\textsubscript{s2} (\%)} \\ \hline
FLAN-Alpaca-base & 10.00 & 8.24 \\ \hline
FLAN-Alpaca-large & 33.53 & 25.29 \\ \hline
FLAN-Alpaca-xl & 0.00 & 0.00 \\ \hline
FLAN-T5-small & 10.59 & 6.47 \\ \hline
FLAN-T5-base & 0.59 & 0.00 \\ \hline
FLAN-T5-large & 2.35 & 0.88 \\ \hline
FLAN-T5-xl & 1.18 & 0.88 \\ \hline
mT0-small & 34.71 & 28.24 \\ \hline
mT0-base & 0.29 & 0.29 \\ \hline
mT0-large & 0.00 & 0.00 \\ \hline
mT0-xl & 0.00 & 0.00 \\ \hline
RoBERTa-dyna-r1 & - & - \\ \hline
RoBERTa-dyna-r2 & - & - \\ \hline
RoBERTa-dyna-r3 & - & - \\ \hline
RoBERTa-dyna-r4 & - & - \\ \hline
GPT-3.5-turbo & 1.47 & 0.88 \\ \hline
GPT-4o & 0.29 & 0.00 \\ \hline
Moderation API & - & - \\ \hline
Perspective API & - & - \\ \hline
DeepSeek LLM-7b & 98.82 & 97.65 \\ \hline
\end{tabular}
\caption{Hallucination rates.`-' if not applicable as models are non-generative.}
\label{tab:hall_rates_details}
\end{table}

\section{Benchmarks Results}
\label{sec:appendix_benchmarks}
We prompt language models on four established hate speech benchmarks for binary hate speech detection using the same instructions as in  \citet{plaza2023respectful-toxic}. 
In Table \ref{tab:f1_hatecheck}, Table \ref{tab:f1_dynabench}, Table \ref{tab:f1_hatexplain}, and Table \ref{tab:f1_implicit_hate}, we report macro F1 scores and the percentage of outputs that followed the instruction as expected for each benchmark. 
Interestingly, DeepSeek LLM shows incredibly low percentages of expected outputs. 
Moreover, we report the Spearman's rank correlation coefficients across static hate speech benchmarks in Table \ref{tab:corr_static_benchs}. 
Overall, the rankings of the models exhibit a positive, non-negligible correlation even though each static benchmark focuses on a specific characteristic of hate speech, namely offensiveness for HateXplain, expressiveness for Implicit Hate, target-based functionality tests for HateCheck, and adversarial examples for Dynabench.
The highest correlation of models' ranking is between Dynabench and HateXplain benchmarks with an average coefficient equal to 0.8647. 
The lowest correlation instead is between HateXplain and Implicit Hate, which however is expected as they measure two very different aspects of hate speech, namely offensiveness and expressiveness \cite{dibonaventura2025detection_explanation_LLM}. 
The average correlation coefficient among all pairs of static evaluations is 0.36 (i.e., \(\frac{1}{6}(0.3865+0.2361+0.3203+0.8647+0.2421+0.0917)\)).

\begin{table}[h!]
\tiny
\centering
\begin{tabular}{|l|c|c|}
\hline
\textbf{Model} & \textbf{Macro F1} & \textbf{Expected Output (\%)} \\ \hline
FLAN-Alpaca-base & .3739 & 95.95 \\ \hline
FLAN-Alpaca-large & .7094 & 100.00 \\ \hline
FLAN-Alpaca-xl & .7348 & 100.00 \\ \hline
FLAN-T5-small & .2322 & 91.34 \\ \hline
FLAN-T5-base & .6023 & 99.97 \\ \hline
FLAN-T5-large & .6909 & 99.30 \\ \hline
FLAN-T5-xl & .7383 & 99.79 \\ \hline
mT0-small & .2747 & 25.78 \\ \hline
mT0-base & .2472 & 99.92 \\ \hline
mT0-large & .6103 & 99.22 \\ \hline
mT0-xl & .4779 & 100.00 \\ \hline
RoBERTa-dyna-r1 & .6235 & 100.00 \\ \hline
RoBERTa-dyna-r2 & .8299 & 100.00 \\ \hline
RoBERTa-dyna-r3 & .9207 & 100.00 \\ \hline
RoBERTa-dyna-r4 & .9485 & 100.00 \\ \hline
GPT-3.5-turbo & .7135 & 99.65 \\ \hline
GPT-4o & .7394 & 100.00 \\ \hline
Moderation API & .5142 & 100.00 \\ \hline
Perspective API & .7489 & 100.00 \\ \hline
DeepSeek LLM-7b & .3750 & 0.54 \\ \hline
\end{tabular}
\caption{Macro F1 and Expected Output rate on HateCheck benchmark.}
\label{tab:f1_hatecheck}
\end{table}

\begin{table}[h!]
\tiny
\centering
\begin{tabular}{|l|c|c|}
\hline
\textbf{Model} & \textbf{Macro F1} & \textbf{Expected Output (\%)} \\ \hline
FLAN-Alpaca-base & .3389 & 87.01 \\ \hline
FLAN-Alpaca-large & .5319 & 99.98 \\ \hline
FLAN-Alpaca-xl & .5744 & 100.00 \\ \hline
FLAN-T5-small & .3067 & 88.90 \\ \hline
FLAN-T5-base & .4971 & 99.57 \\ \hline
FLAN-T5-large & .5220 & 98.58 \\ \hline
FLAN-T5-xl & .5855 & 99.73 \\ \hline
mT0-small & .3215 & 68.65 \\ \hline
mT0-base & .3309 & 99.03 \\ \hline
mT0-large & .5252 & 99.18 \\ \hline
mT0-xl & .4381 & 100.00 \\ \hline
RoBERTa-dyna-r1 & .5829 & 100.00 \\ \hline
RoBERTa-dyna-r2 & .7022 & 100.00 \\ \hline
RoBERTa-dyna-r3 & .8120 & 100.00 \\ \hline
RoBERTa-dyna-r4 & .8104 & 100.00 \\ \hline
GPT-3.5-turbo & .5045 & 99.48 \\ \hline
GPT-4o & .5728 & 99.47 \\ \hline
Moderation API & .4219 & 99.96 \\ \hline
Perspective API & .5255 & 100.00 \\ \hline
DeepSeek LLM-7b & .4203 & 7.86 \\ \hline
\end{tabular}
\caption{Macro F1 and Expected Output rate on Dynabench benchmark.}
\label{tab:f1_dynabench}
\end{table}

\begin{table}[h!]
\tiny
\centering
\begin{tabular}{|l|c|c|}
\hline
\textbf{Model} & \textbf{Macro F1} & \textbf{Expected Output (\%)} \\ \hline
FLAN-Alpaca-base & .4333 & 93.34 \\ \hline
FLAN-Alpaca-large & .6015 & 100.00 \\ \hline
FLAN-Alpaca-xl & .6827 & 100.00 \\ \hline
FLAN-T5-small & .2895 & 98.34 \\ \hline
FLAN-T5-base & .5704 & 99.69 \\ \hline
FLAN-T5-large & .5479 & 99.01 \\ \hline
FLAN-T5-xl & .7201 & 100.00 \\ \hline
mT0-small & .2844 & 65.23 \\ \hline
mT0-base & .3419 & 98.23 \\ \hline
mT0-large & .4928 & 99.48 \\ \hline
mT0-xl & .4829 & 100.00 \\ \hline
RoBERTa-dyna-r1 & .6989 & 100.00 \\ \hline
RoBERTa-dyna-r2 & .6989 & 100.00 \\ \hline
RoBERTa-dyna-r3 & .7096 & 100.00 \\ \hline
RoBERTa-dyna-r4 & .7077 & 100.00 \\ \hline
GPT-3.5-turbo & .4539 & 99.58 \\ \hline
GPT-4o & .5732 & 99.38 \\ \hline
Moderation API & .5055 & 100.00 \\ \hline
Perspective API & .6621 & 100.00 \\ \hline
DeepSeek LLM-7b & .4266 & 10.01 \\ \hline
\end{tabular}
\caption{Macro F1 and Expected Output rate on HateXplain benchmark.}
\label{tab:f1_hatexplain}
\end{table}

\begin{table}[h!]
\tiny
\centering
\begin{tabular}{|l|c|c|}
\hline
\textbf{Model} & \textbf{Macro F1} & \textbf{Expected Output (\%)} \\ \hline
FLAN-Alpaca-base & .4091 & 93.53 \\ \hline
FLAN-Alpaca-large & .5625 & 100.00 \\ \hline
FLAN-Alpaca-xl & .6167 & 100.00 \\ \hline
FLAN-T5-small & .3870 & 97.49 \\ \hline
FLAN-T5-base & .5334 & 99.30 \\ \hline
FLAN-T5-large & .4995 & 99.12 \\ \hline
FLAN-T5-xl & .6215 & 100.00 \\ \hline
mT0-small & .3896 & 47.25 \\ \hline
mT0-base & .4022 & 94.09 \\ \hline
mT0-large & .4673 & 96.65 \\ \hline
mT0-xl & .4073 & 100.00 \\ \hline
RoBERTa-dyna-r1 & .6146 & 100.00 \\ \hline
RoBERTa-dyna-r2 & .6377 & 100.00 \\ \hline
RoBERTa-dyna-r3 & .6184 & 100.00 \\ \hline
RoBERTa-dyna-r4 & .6491 & 100.00 \\ \hline
GPT-3.5-turbo & .3718 & 99.39 \\ \hline
GPT-4o & .4815 & 99.58 \\ \hline
Moderation API & .4009 & 99.95 \\ \hline
Perspective API & .6017 & 100.00 \\ \hline
DeepSeek LLM-7b & .4590 & 2.75 \\ \hline
\end{tabular}
\caption{Macro F1 and Expected Output rate on Implicit Hate benchmark.}
\label{tab:f1_implicit_hate}
\end{table}

\begin{table}[h!]
\tiny
\centering
\begin{tabular}{|l|c|c|c|c|}
\hline
\textbf{} & \textbf{HateCheck} & \textbf{Dynabench} & \textbf{HateXplain} & \textbf{Implicit Hate} \\ \hline

\textbf{HateCheck} & 1. & 0.3865 & 0.2361 & 0.3203 \\ \hline
\textbf{Dynabench} & - & 1. & 0.8647 & 0.2421 \\ \hline
\textbf{HateXplain} & - & - & 1. & 0.0917 \\ \hline
\textbf{Implicit Hate} & - & - & - & 1. \\ \hline
\end{tabular}
\caption{Spearman's rank correlation coefficient across static hate speech benchmarks.}
\label{tab:corr_static_benchs}
\end{table}

\section{Correlation Analysis}
\label{sec:appendix_confidence}

We use the Spearman’s rank correlation to measure the strength and direction of association between static and time-sensitive evaluations. 
The Spearman's rank correlation coefficient can take a value from +1 to -1 where a value of +1 means a perfect positive correlation, a value of 0 means no correlation, and a value of -1 means a perfect negative association of rank. 
In addition to the correlation coefficients shown in Table \ref{tab:corr_rankings} of the main paper, we report their confidence intervals in Table \ref{tab:confidence_rankings} below. 
These confidence intervals \((c_{lower}, c_{upper})\) are computed as follows. 

\[ c_{lower} = \frac{e^{2L}-1}{e^{2L}+1} \]

\[ c_{upper} = \frac{e^{2U}-1}{e^{2U}+1} \]

where 

\[ L =  Z - \frac{Z_{1-\alpha/2}}{\sqrt{n-3}} \]

\[ U =  Z + \frac{Z_{1-\alpha/2}}{\sqrt{n-3}} \]

\[ Z = \frac{1}{2} \ln(\frac{1+\rho}{1-\rho}) \]

with significance level \(\alpha=0.10\), sample size \(n=20\), and Spearman's rank correlation coefficient \(\rho\) being the ones in Table \ref{tab:corr_rankings}. 
The results can be interpreted as there is a 90\% chance that the confidence intervals shown below contain the true population correlation coefficient between static and time-sensitive evaluations of language models.
Overall, these intervals suggest a negative or negligible correlation between static and time-sensitive rankings, with a skewed tendency toward negative correlations. 
Note that sample size affects this estimate and that a larger sample could provide a more precise assessment. 

Moreover, we report the confidence intervals of the correlation coefficients of models' ranking among static evaluations in Table \ref{tab:confidence_rankings_static}.

\begin{table}[h!]
\tiny
\centering
\begin{tabular}{|l|c|c|}
\hline
\textbf{$\downarrow$ Static / Time-sensitive $\rightarrow$} & \textbf{Experiment 1} & \textbf{Experiment 2} \\ \hline
\textbf{HateCheck} & (-0.586, 0.126) & (-0.438, 0.317) \\ \hline
\textbf{Dynabench} & (-0.504, 0.238) & (-0.613, 0.083) \\ \hline
\textbf{HateXplain} & (-0.578, 0.138) & (-0.528, 0.207) \\ \hline
\textbf{Implicit Hate} & (-0.597, 0.110) & (-0.203, 0.532) \\ \hline
\end{tabular}
\caption{Confidence intervals of Spearman's rank correlation coefficient between static and time-sensitive evaluations.}
\label{tab:confidence_rankings}
\end{table}

\begin{table}[h!]
\tiny
\centering
\begin{tabular}{|l|c|c|c|}
\hline
\textbf{$\downarrow$ Static / Static $\rightarrow$} & \textbf{Dynabench} & \textbf{HateXplain} & \textbf{Implicit Hate} \\ \hline

\textbf{HateCheck} & (0.009, 0.668) & (-0.157, 0.565) & (-0.067, 0.624)\\ \hline
\textbf{Dynabench} & - & (0.722, 0.937) & (-0.151, 0.569) \\ \hline
\textbf{HateXplain} & - & - & (-0.298, 0.455) \\ \hline
\textbf{Implicit Hate} & - & - & - \\ \hline
\end{tabular}
\caption{Confidence intervals of Spearman's rank correlation coefficient between static evaluations.}
\label{tab:confidence_rankings_static}
\end{table}

\end{document}